\documentclass[10pt,twocolumn,letterpaper]{article}

\usepackage{cvpr}
\usepackage{times}
\usepackage{epsfig}
\usepackage{graphicx}
\usepackage{amsmath}
\usepackage{amssymb}
\usepackage{multirow}
\usepackage{color}
\usepackage{booktabs}

\usepackage[pagebackref=true,breaklinks=true,letterpaper=true,colorlinks,citecolor=black,linkcolor=red,bookmarks=false]{hyperref}
\usepackage{animate}
\cvprfinalcopy 

\def\cvprPaperID{7166} 

\ifcvprfinal\fi
\begin{document}

\title{Deep Blind Video Super-resolution}

\vspace{-8mm}
\author{Jinshan Pan$^{1}$\quad Songsheng Cheng$^{1}$\quad Jiawei Zhang$^{2}$ \quad Jinhui Tang$^{1}$\\
$^{1}$Nanjing University of Science and Technology \quad $^{2}$SenseTime Research \\
\\
}

\maketitle

\begin{figure}[t]\footnotesize
\vspace{-3.16in}
\begin{minipage}{\textwidth}
\begin{center}
\begin{tabular}{cccccccc}
\multicolumn{3}{c}{\multirow{5}*[68pt]{\includegraphics[width=0.368\linewidth, height = 0.326\linewidth]{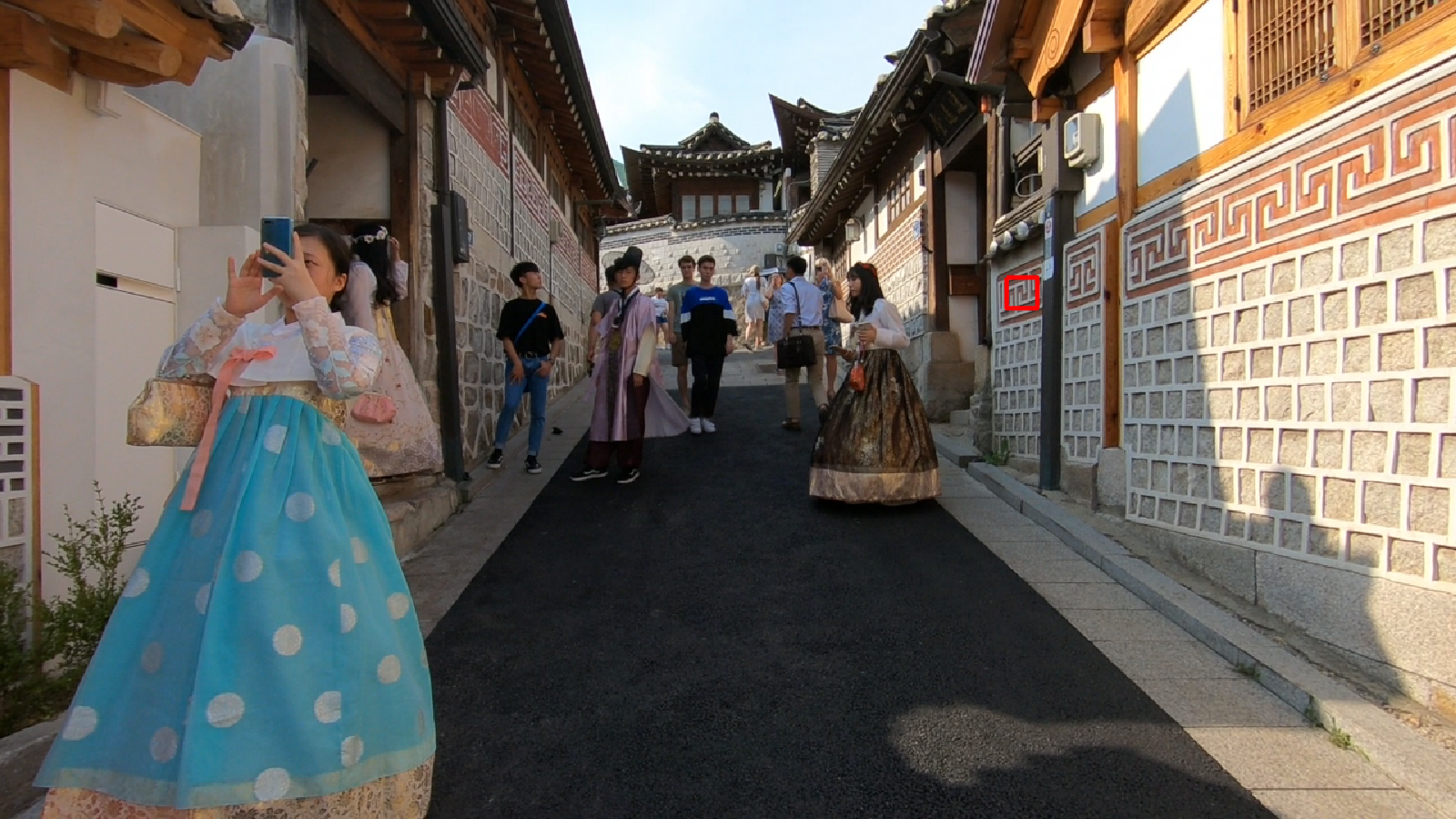}}}&\hspace{-4.0mm}
\includegraphics[width=0.15\linewidth]{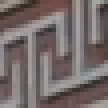} &\hspace{-4.0mm}
\includegraphics[width=0.15\linewidth]{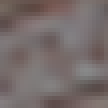} &\hspace{-4.0mm}
\includegraphics[width=0.15\linewidth]{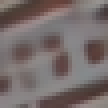} &\hspace{-4.0mm}
\includegraphics[width=0.15\linewidth]{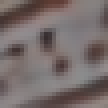} \\
\multicolumn{3}{c}{~} &\hspace{-4.5mm}  (b) HR patch &\hspace{-4.5mm}  (c) Bicubic &\hspace{-4.5mm}  (d) RCAN~\cite{RCAN}  &\hspace{-4.5mm}  (e) DUF~\cite{jo/cvpr18} \\
%
\multicolumn{3}{c}{~} & \hspace{-4.5mm}
\includegraphics[width=0.15\linewidth]{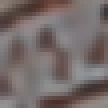} & \hspace{-4.0mm}
\includegraphics[width=0.15\linewidth]{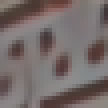} & \hspace{-4.0mm}
\includegraphics[width=0.15\linewidth]{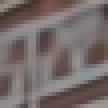} & \hspace{-4.0mm}
\includegraphics[width=0.15\linewidth]{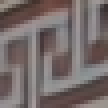} \\
\multicolumn{3}{c}{\hspace{-4.5mm} (a) Ground truth high-resolution (HR) image} &  \hspace{-4.5mm} (f) TOF~\cite{tof} &\hspace{-4.5mm}  (g) RBPN~\cite{VDBPN/cvpr19} &\hspace{-4.5mm}  (h) w/o kernel modeling & \hspace{-4.5mm} (i) Ours\\
\end{tabular}
\end{center}
\vspace{-3mm}
\caption{Video super-resolution result ($\times 4$). Existing video super-resolution algorithms usually assume the blur kernel in the degradation is known and do not model the blur kernel in the restoration process. We show that without modeling the blur kernel does not effectively capture the intrinsic characteristics of the video super-resolution problem which thus leads to over-smoothed results (see (c)-(h)). Our algorithm jointly estimates blur kernels, motion fields, and latent frames, which is able to generate clearer results and better recovers structural details.
}%
\label{fig: teaser}
\end{minipage}
\vspace{-6mm}
\end{figure}

\begin{abstract}

   %
   Existing video super-resolution (SR) algorithms usually assume that the blur kernels in the degradation process are known and do not model the blur kernels in the restoration.
   However, this assumption does not hold for video SR and usually leads to over-smoothed super-resolved images.
   In this paper, we propose a deep convolutional neural network (CNN) model to solve video SR by a blur kernel modeling approach.
   The proposed deep CNN model consists of motion blur estimation, motion estimation, and latent image restoration modules.
   The motion blur estimation module is used to provide reliable blur kernels.
   With the estimated blur kernel, we develop an image deconvolution method based on the image formation model of video SR to generate intermediate latent images
   so that some sharp image contents can be restored well.
   However, the generated intermediate latent images may contain artifacts.
   To generate high-quality images, we use the motion estimation module to explore the information from adjacent frames, where the motion estimation can constrain the deep CNN model for better image restoration.
   %
   %
   We show that the proposed algorithm is able to generate clearer images with finer structural details.
   \vspace{4mm}
   Extensive experimental results show that the proposed algorithm performs favorably against state-of-the-art methods.
\end{abstract}

\vspace{-2mm}
\section{Introduction}
\label{sec: introduction}

Video super-resolution (SR) aims to estimate high-resolution (HR) frames from a low-resolution (LR) sequence.
It is a fundamental problem in the vision and graphics communities and has received active research efforts within the last decade as high-definition devices have been widely used in our daily lives.
As the HR sequences are usually contaminated by unknown blur, it is quite challenging to restore HR images from low-resolution sequences.

Since video SR is an ill-posed problem, conventional methods usually estimate underlying motion and latent images simultaneously in
a variational approach~\cite{Milanfar/tip04,prior/eccv96,prior/tip01,DBLP:conf/cvpr/ShaharFI11,Bayesian/vsr/tpami14,maziyang/vsr/cvpr15}.
To improve the performance, kinds of hand-crafted priors on the latent images and motion fields have been widely used in these methods.
In spite of achieving decent results, these algorithms usually need to complex energy functions or complex matching processes and the performance is limited by the hand-crafted priors.
In addition, most of these algorithms usually use known blur kernels (e.g., Gaussian blur kernel, Bicubic kernel) and do not model blur kernels in the restoration, which cannot effectively capture the intrinsic characteristics of video SR~\cite{Bayesian/vsr/tpami14}.

%

%
Motivated by the first end-to-end trainable network for single image SR~\cite{SRCNN/tpmai}, lots of methods based on deep convolutional neural networks (CNNs) have been proposed~\cite{VDSR,Accelerating/dong,DBPN,RCAN,edsr,SRGAN}.
These algorithms achieve decent results in single image SR. However, directly using these algorithms cannot solve the video SR problem well.
To overcome this problem, most existing algorithms focus on developing effective motion fields and alignment estimation methods.
For example, the subpixel motion compensation based on optical flow~\cite{xintao/iccv17}, deformable alignment networks~\cite{TDAN,edvr}, and spatial alignment networks~\cite{liu/iccv17,vespcn,tof}.
%
%
%
To better restore latent images, the recurrent approaches and Generative Adversarial Networks (GANs) have been developed~\cite{TecoGAN,GAN/video/tip19}.
These methods significantly promote the progress of video SR.
However, they usually assume the blur kernel is known (e.g., Bicubic kernel). Therefore, without modeling the blur kernel usually leads to over-smoothed results (Figure~\ref{fig: teaser}).

To overcome this problem, several algorithms explicitly estimate blur kernels for SR~\cite{Tomer/blindsr/iccv13,Gu/cvpr19,Irani/cvpr18,iGAN/kernel/nips19}.
These algorithms show that using the estimated blur kernels for image SR is able to improve the results significantly~\cite{Tomer/blindsr/iccv13,iGAN/kernel/nips19}.
However, these algorithms are mainly developed for single image SR which cannot be extended to video SR directly.
The methods by~\cite{Bayesian/vsr/tpami14,maziyang/vsr/cvpr15} simultaneously estimate underlying motion and blur kernels for image restoration. However, the performance is limited by the hand-crafted image priors.

To overcome the above problems, we propose an effective video SR algorithm that simultaneously estimates underlying motion blur, motion field, and latent image by deep CNN models so that our method can not only avoid the hand-crafted priors and but also effectively estimate blur kernels and motion fields for better image restoration.
%
The proposed algorithm mainly consists of motion blur estimation, motion field estimation, and latent image restoration modules.
The motion blur estimation generates blur kernels based on the image formation of video SR and is able to provide intermediate latent images with sharp contents.
The motion field estimation is used to explore the spatio-temporal information from adjacent frames so that it can guide the deep CNN model for better image restoration.
By training the proposed algorithm in an end-to-end manner, it is able to generate clearer images with finer structural details (Figure~\ref{fig: teaser}).

The main contributions are summarized as follows:
\begin{itemize}
  \item We propose an effective video SR algorithm that simultaneously estimates blur kernels, motion fileds, and latent images by deep CNN models.
  \item We develop an effective kernel estimation method and image deconvolution algorithm based on the image formation of video SR.
  To restore high-quality images, we explore the spatio-temporal information from the adjacent frames so that it can guide the deep CNN model for better image restoration.
  \item We both quantitatively and qualitatively evaluate the proposed algorithm on benchmark datasets and real-world videos and show that it performs favorably against state-of-the-art methods.
\end{itemize}

\section{Related Work}
\label{sec: Related-Work}
We briefly discuss methods most relevant to this work and put this work in proper context.
\vspace{-2mm}
{\flushleft \bf{Variational approach.}}
Since video SR is highly ill-posed, early approaches mainly focus on developing effective priors~\cite{Milanfar/tip04,prior/eccv96,prior/tip01,DBLP:conf/cvpr/ShaharFI11} on the HR images to solve this problem.
As these methods usually use known blur kernels to approximate the real ones which will lead to over-smoothed results. Several methods~\cite{Bayesian/vsr/tpami14,maziyang/vsr/cvpr15} simultaneously estimate motion fileds, blur kernels, and latent images in a Maximum a posteriori (MAP) framework.
In~\cite{Bayesian/vsr/tpami14}, Liu and Sun solve video SR by a Bayesian framework, where the motion fileds, blur kernels, latent images, and noise levels are estimated simultaneously.
Ma et al.~\cite{maziyang/vsr/cvpr15} propose an effective Expectation Maximization (EM) framework to jointly solve video SR and blur estimation.
Although promising results have been achieved, these algorithms require solving complex optimization problems. In addition, the performance is limited by the hand-crafted priors.
\vspace{-2mm}
{\flushleft \bf{Deep learning approach.}}
Motivated by the success of deep learning-based single image SR~\cite{SRCNN/tpmai,VDSR,Accelerating/dong,DBPN,RCAN,edsr,SRGAN},
several methods~\cite{Huang/nips15,liao/iccv15,Kappeler/tci16,vespcn,liu/iccv17,xintao/iccv17,tof,jo/cvpr18,VDBPN/cvpr19,edvr,TDAN} explore the spatio-temporal information for video SR.
Huang et al.~\cite{Huang/nips15} develop an effective bidirectional recurrent convolutional network to model the long-term contextual information.
Some algorithms~\cite{liao/iccv15,Kappeler/tci16} first estimate motion fields based on the hand-crafted priors and then use a deep CNN model to restore high-quality images.
In~\cite{vespcn}, Caballero et al. develop an effective motion compensation and explore the spatio-temporal information for video SR.
Liu et al.~\cite{liu/iccv17} develop a temporal adaptive neural network and a spatial alignment network to better explore the temporal information.
In~\cite{xintao/iccv17}, Tao et al. propose an effective subpixel motion compensation layer based on the estimated motion fields for video SR.
Xue et al.~\cite{tof} demonstrate the effect of optical flow on video image restoration and propose an effective video restoration framework to solve general video restoration problems.
Instead of explicitly using optical flow for alignment, Jo et al.~\cite{jo/cvpr18} dynamically estimate upsampling filters.
In~\cite{VDBPN/cvpr19}, Haris et al. extend the deep back-projection method~\cite{DBPN} by a recurrent network.
Wang et al.~\cite{edvr} improve the deformable convolution~\cite{TDAN} and develop an effective temporal and spatial attention to solve video restoration.
This algorithm wins the champions in the NTIRE19 video restoration~\cite{REDS}.
To generate more realistic images, GANs have been used to solve the both single~\cite{SRGAN,Sajjadi_2017_ICCV,GAN/data/eccv18}
and video~\cite{TecoGAN,GAN/video/tip19} SR problems.
These algorithms generate decent results on video SR. However, these algorithms either explicitly or implicitly assume that the blur kernels are known and do not model the blur kernels for SR, which accordingly leads to over-smoothed results.

Estimating blur kernels has been demonstrated effective for image SR, especially for the details restoration~\cite{Tomer/blindsr/iccv13,Gu/cvpr19,Irani/cvpr18,iGAN/kernel/nips19,blindsr/sa09,zhangkai/cvpr19/sr}.
However, these algorithms are designed for single image SR. Few of them have been developed for video SR.
Different from these methods, we propose a deep CNN model to simultaneously estimate blur kernels, motion fields, and latent frames so that high-quality videos can be better-restored.
%

\begin{figure*}[!t]\footnotesize
\begin{center}
\begin{tabular}{c}
\includegraphics[width = 0.98\linewidth]{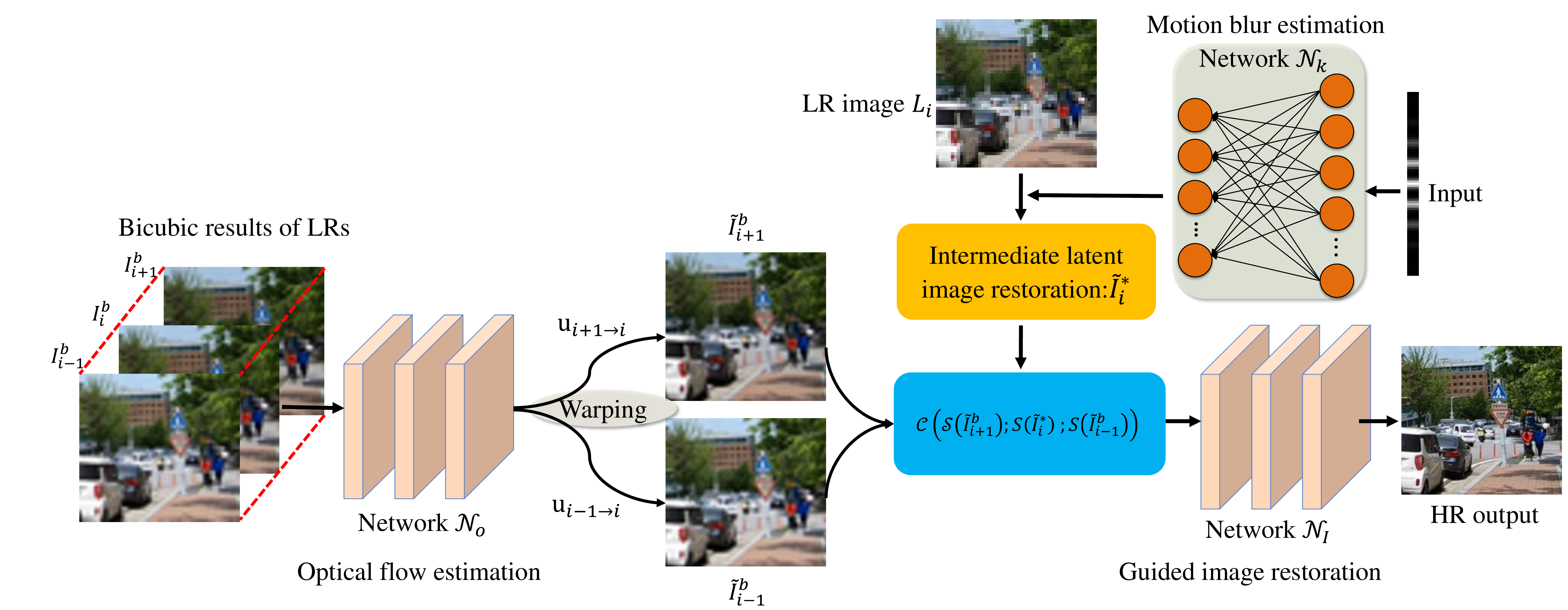}\\
\end{tabular}
\end{center}
\vspace{-4mm}
\caption{An overview of the proposed method. The proposed algorithm takes three adjacent frames and initialized input kernel as the input and super-resolves the center image (i.e., $L_i$).
First, we use $\mathcal{N}_k$ to estimate blur kernels from initialized input kernel, where the Gaussian blur kernel is used as the initialized kernel. Then, we generate an intermediate HR image ($\tilde{I}_{i}^{*}$) based on an image deconvolution method with the estimated blur kernels.
To remove the artifacts in $\tilde{I}_{i}^{*}$, we compute the optical flow based on the Bicubic usampling results of three adjacent frames and generate the warped images (i.e., $\tilde{I}_{i+1}^{b}$, $\tilde{I}_{i-1}^{b}$) to guide the restoration of $\tilde{I}_{i}^{*}$ based on $\mathcal{N}_I$.
The proposed algorithm is jointly trained in an end-to-end manner and generates better high-quality images.
The mathematical operators are detailed in main contents.
}
\label{fig: flow-chart}
\vspace{-4mm}
\end{figure*}

\section{Revisiting Variational Methods}
\label{sec: motivation}
The proposed algorithm is motivated by the variational methods~\cite{Milanfar/tip04,Bayesian/vsr/tpami14,maziyang/vsr/cvpr15} for video SR.
In this section, we first revisit how these variational methods~\cite{Milanfar/tip04,Bayesian/vsr/tpami14,maziyang/vsr/cvpr15} solve video SR and then introduce the proposed algorithm.

Following the definitions of~\cite{Milanfar/tip04,Bayesian/vsr/tpami14,maziyang/vsr/cvpr15}, the degradation model for video SR is:
\begin{equation}
L_j = \mathbf{SK}\mathbf{F}_{\mathrm{u}_{i\to j}}I_i + n_j,
\label{eq: video-sr-model}
\end{equation}
where $\{L_{j}\}_{j =i-N}^{i+N}$ denote a set of LR images with $2N+1$ frames; $I_i$ denotes the HR image; $n_j$ denotes image noise,
$\mathbf{S}$ and $\mathbf{K}$ denote the matrix form of down-sampling and blur kernel; $\mathbf{F}_{\mathrm{u}_{i\to j}}$ denotes the warping matrix w.r.t. optical flow $\mathrm{u}_{i\to j}$,
and $\mathrm{u}_{i\to j}$ denotes the optical flow from $I_i$ to $I_j$.

Based on the degradation model~\eqref{eq: video-sr-model}, the HR image ${I_i}$, optical flow $\mathrm{u}_{i\to j}$, and blur kernel ${K}$ can be estimated by a Maximum a posteriori (MAP):
\begin{equation}
\begin{split}
\{I_i^*, K^*, \{\mathrm{u}^*_{i\to j}\}\} &= \arg\max_{{I_i}, {K}, \{\mathrm{u}_{i\to j}\}}p({I_i}, {K}, \{\mathrm{u}_{i\to j}\}|\{{L}_j\}), \\
& = p({I_i})p({K})\prod_jp(\mathrm{u}_{i\to j}) p({{L}_i}|{I_i}, {K})\\
&\qquad\qquad   \prod_{j\neq i}p(\{{L}_j\}|{I_i}, {K}, \{\mathrm{u}_{i\to j}\})
\label{eq: video-sr-model-map}
\end{split}
\end{equation}
Using hand-crafted image priors $\rho({I_i})$, $\varphi(\mathrm{u}_{i\to j})$, and $\phi(K)$ on the HR image ${I_i}$, optical flow $\mathrm{u}_{i\to j}$, and blur kernel ${K}$, respectively, the video SR process can be achieved by alternatively minimizing~\cite{Bayesian/vsr/tpami14}:
\begin{equation}
\begin{split}
I_i^* = &\arg\min_{I_i}\|\mathbf{SK}I_i - L_i\| + \\
&\sum_{j=i-N, j\neq i}^{i+N}\|\mathbf{SK}\mathbf{F}_{\mathrm{u}_{i\to j}}I_i - L_j\| + \rho({I_i}),
\label{eq: video-sr-model-latent-frame}
\end{split}
\end{equation}
\begin{equation}
\begin{split}
\mathrm{u}_{i\to j}^* = \arg\min_{\mathrm{u}_{i\to j}}\|\mathbf{SK}\mathbf{F}_{\mathrm{u}_{i\to j}}I_i - L_j\| + \varphi(\mathrm{u}_{i\to j}),
\label{eq: video-sr-model-optical-flow}
\end{split}
\end{equation}
and
\begin{equation}
\begin{split}
K^* = \arg\min_K\|\mathbf{S}\mathbf{T}_{I_i}K - L_i\| + \phi(K),
\label{eq: video-sr-model-kernel}
\end{split}
\end{equation}
where $\mathbf{T}_{I_i}$ is a matrix of latent HR image $I_i$ w.r.t. $K$~\cite{Bayesian/vsr/tpami14}.

Although the video SR algorithms~\cite{Bayesian/vsr/tpami14,maziyang/vsr/cvpr15} based on above model have been demonstrated effective in both benchmark datasets and real-world videos,
they need to define the hand-crafted image priors $\rho({I_i})$, $\varphi(\mathrm{u}_{i\to j})$, and $\phi(K)$ which usually lead to highly non-convex objective function~\eqref{eq: video-sr-model-map}.
This makes the video SR problem more difficult to solve. In addition, the performance of video SR is limited by the hand-crafted image priors.
We further note that most existing deep learning-based methods usually employ deep CNN models to solve video SR problem.
Although these methods do not need to define hand-crafted priors, they cannot capture the intrinsic characteristics of video SR as the blur kernel is assumed to be known (e.g., Bicubic~\cite{edvr}, Gaussian~\cite{gaussian/kernel/cvpr18}) and do not model it in the SR process, which accordingly lead to over-smoothed results.

To overcome these problems, we develop an effective deep CNN model which consists of motion blur estimation, motion field estimation, and latent image restoration for video SR.
The proposed model does not need the hand-crafted priors and is able to capture the intrinsic characteristics of degradation process in video SR by modeling blur kernels. Thus, it can generate much better super-resolved videos with clearer structural details.

\section{Proposed Algorithm}
%
%
The overview of the proposed method is shown in Figure~\ref{fig: flow-chart}.
In the following, we explain the main ideas for each component in details.

\subsection{Motion blur estimation}
\label{ssec: Optical flow estimation}
%
We note that the motion blur estimation based on~\eqref{eq: video-sr-model-kernel} needs to define a hand-crafted prior $\phi(K)$ which usually leads to a complex optimization process.
To avoid hand-crafted priors and the complex optimization process, we develop a deep CNN model $\mathcal{N}_k$ to effectively estimate motion blur kernels.
The network $\mathcal{N}_k$ takes initialized Gaussian kernels as input and refine it based on the degradation model~\eqref{eq: video-sr-model}.
Given the HR images $\{I_i\}$ and the corresponding LR images $\{L_i\}$, we use the first term of~\eqref{eq: video-sr-model-kernel}
(which is related to the degradation model~\eqref{eq: video-sr-model}) to constrain the deep CNN model $\mathcal{N}_k$:
\begin{equation}
\mathcal{L}_k = \|\mathbf{S}\tilde{K}I_i - L_i\|_1,
\label{eq: kernel-estimation-loss}
\end{equation}
where $\tilde{K}$ denotes the output of the deep CNN model $\mathcal{N}_k$ and $\ell_1$ norm is used.
Similar to~\cite{dongwei/deepdeblur}, the motion blur estimation network $\mathcal{N}_k$ consists of two fully connected layers, where the first fully connected layer is followed by a ReLU activation function
and the second one is follow by a Softmax function to ensure that each element of the blur kernel is nonnegative and the summation of all elements is $1$.

\begin{figure}[!t]\footnotesize
\centering
\begin{tabular}{ccc}
\includegraphics[width=0.32\linewidth]{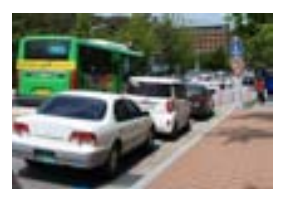} &\hspace{-4mm}
\includegraphics[width=0.32\linewidth]{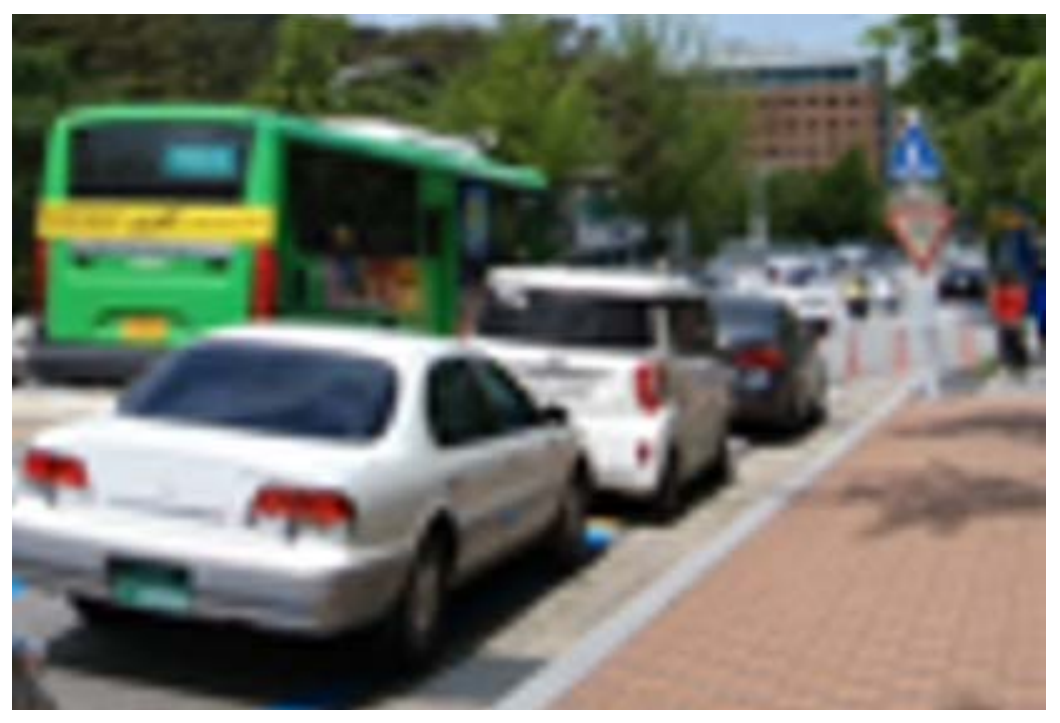} &\hspace{-4mm}
\includegraphics[width=0.32\linewidth]{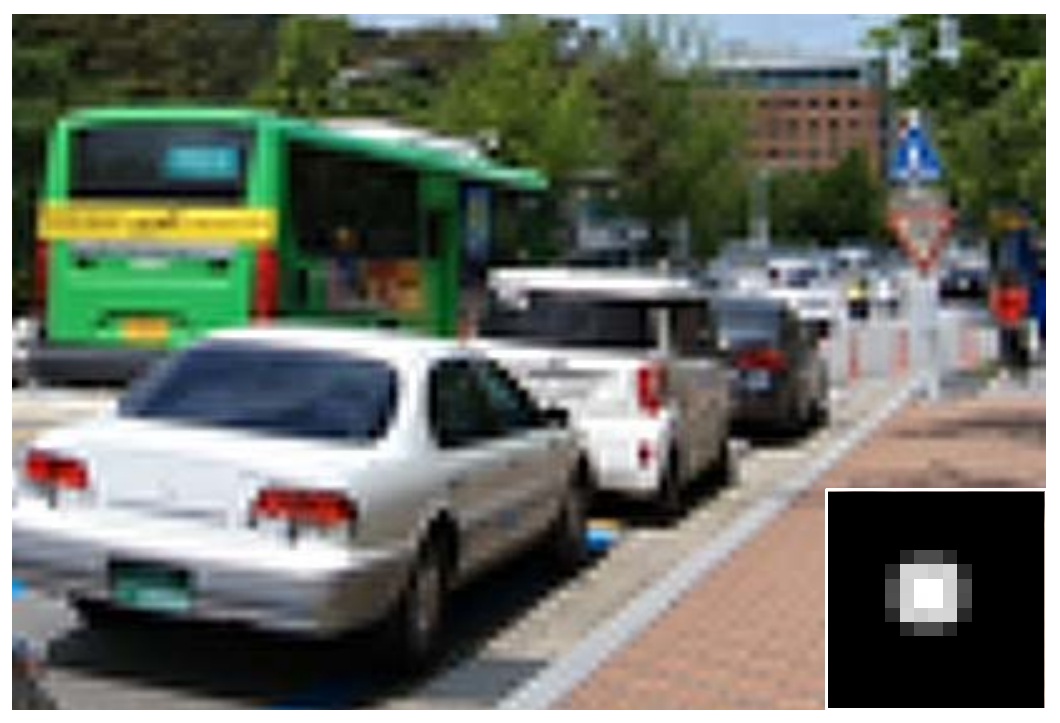} \\
(a) LR &\hspace{-4mm}  (b) Bicubic &\hspace{-4mm}  (c) Deblurred \& kernel\\
\end{tabular}
\caption{Effect of the intermediate latent image restoration and motion blur estimation. Using the estimated blur kernel to deblur LR images generates sharper images (c).}%
\label{fig: intermediate-result-kernel}
\vspace{-4mm}
\end{figure}

Figure~\ref{fig: intermediate-result-kernel}(c) shows the estimated blur kernel from bicubic downsampling LR images. We note that the shape of the blur kernel is quite similar to that of Bicubic blur kernel~\cite{levin/etal/blindsr}. We will demonstrate the effectiveness of the motion blur estimation in Section~\ref{sec: Analysis and Discussions}.

\subsection{Intermediate latent image restoration}
\label{sec: Latent frame restoration}
%

With the blur kernel $K$, we can estimate HR image from input LR image $L_i$ according to~\eqref{eq: video-sr-model-latent-frame}.
However, solving~\eqref{eq: video-sr-model-latent-frame} needs the optical flow and image prior.
Recent algorithms~\cite{jiawei/cvpr17/deblur,kaizhang/cvpr17/deblur} show that the image prior can be efficiently learned by deep CNN models
so that the restoration process can be achieved by alternatively solving a simple model to restore intermediate latent images and using deep CNN models to remove the noise and artifacts in the intermediate latent images.
Motivated by the success of~\cite{jiawei/cvpr17/deblur,kaizhang/cvpr17/deblur}, we first estimate an intermediate HR image by a simple image deconvolution model and then explore the information of adjacent frames and deep CNN models to restore high-quality images.

To obtain the intermediate HR image efficiently, we propose an image deconvolution model based on the image formation~\eqref{eq: video-sr-model} by:
\begin{equation}
\vspace{-1mm}
\tilde{I}_i^* = \arg\min_{I_i}\|\mathbf{S}\mathbf{\tilde{K}}I_i - L_i\|^2 + \gamma\|\nabla I_i\|^2,
\label{eq: deconvolution-step-1}
\vspace{-1mm}
\end{equation}
where $\|\nabla I_i\|^2$ is used to make the problem well-posed and $L_2$ norm is used to make the problem be efficiently solved, and $\nabla$ denotes the gradient operator.
Note that~\eqref{eq: deconvolution-step-1} is a least square problem. We can get the closed-form solution by:
\begin{equation}
\vspace{-1mm}
\tilde{I}_i^* = \left(\mathbf{\tilde{K}}^{\top}\mathbf{S}^{\top}\mathbf{S}\mathbf{\tilde{K}}^{\top} + \gamma (\mathbf{D}_v^{\top}\mathbf{D}_v + \mathbf{D}_h^{\top}\mathbf{D}_h)\right)^{-1}{\mathbf{\tilde{K}}^{\top}\mathbf{S}^{\top}L_i},
\label{eq: deconvolution-step-1}
\end{equation}
where $\mathbf{D}_h$ and $\mathbf{D}_v$ denote the matrices of derivative filters in horizontal and vertical directions.

Figure~\ref{fig: intermediate-result-kernel}(c) shows the estimated intermediate HR image $\tilde{I}_i^*$.
Note that though $\tilde{I}_i^*$ contains noise and artifacts, it also contains some clear contents which facilitate the following image restoration, especially for the structural details restoration (Figure~\ref{fig: teaser}(i)).
In the following, we will use the adjacent frames and deep CNN models to remove noise and artifacts in $\tilde{I}_i^*$.
\subsection{Optical flow estimation}
\label{ssec: Optical flow estimation}
The optical flow is used to warp adjacent frames to the reference frame and provide more reliable information for the reference frame restoration.
In this work, we use the PWC-Net~\cite{pwcnent/deqing} as the proposed optical flow estimation algorithm given its small model size and decent performance.
We note that the intermediate HR image contains artifacts and noise which may interfere the optical flow estimation.
Thus, we use the Bicubic upsampling result of each LR image to compute the initial optical flow.

Given any three adjacent frames $L_{i-1}$, $L_{i}$, and $L_{i+1}$, we first use the Bicubic upsampling to obtain $I_{i-1}^{b}$, $I_{i}^{b}$, and $I_{i+1}^{b}$, respectively. Then, the PWC-Net (denoted as $\mathcal{N}_o$ in Figure~\ref{fig: flow-chart}) is used to compute optical flow $\mathrm{u}_{i-1\to i}$ and $\mathrm{u}_{i+1\to i}$ based on the Bicubic upsampling results, where the PWC-Net for the computations of $u_{i-1\to i}$ and $u_{i+1\to i}$ shares the same parameters.
Based on the estimated optical flow, we use the bilinear interpolation method to obtain the warped images $I_{i-1}^{b}(\mathrm{x} + \mathrm{u}_{i-1\to i})$
and $I_{i+1}^{b}(\mathrm{x} + \mathrm{u}_{i+1\to i})$ according to~\cite{pwcnent/deqing} (i.e., $\tilde{I}_{i+1}^{b}$, $\tilde{I}_{i-1}^{b}$ in Figure~\ref{fig: flow-chart}).
%
%


\subsection{Guided image restoration}
\label{sec: High-quality HR image restoration}
Using the warped images $\tilde{I}_{i-1}^{b}$ and $\tilde{I}_{i+1}^{b}$ as the guidance,
we can employ existing deep CNN models for image restoration to estimate a high-quality image from $\tilde{I}_i^*$.
In this paper, we use the deep CNN model by~\cite{RCAN} to restore high-quality images,
where we change the network input as the concatenation of $\tilde{I}_{i-1}^{b}$, $\tilde{I}_{i+1}^{b}$, and $\tilde{I}_i^{*}$.
However, as $\tilde{I}_{i-1}^{b}$, $\tilde{I}_{i+1}^{b}$, and $\tilde{I}_i^{*}$ are HR images, they will increase the computational cost.
To overcome this problem, we adopt the space-to-depth transformation~\cite{gaussian/kernel/cvpr18} to divide these HR images into LR ones.
Thus, the high-quality image can be obtained by
\begin{equation}
I_i^* = \mathcal{N}_{I}(\mathcal{C}(\mathcal{S}(\tilde{I}_{i+1}^{b}; \mathcal{S}(\tilde{I}_i^*); \mathcal{S}(\tilde{I}_{i-1}^{b}))),
\label{eq: image-restoration-final}
\end{equation}
where $\mathcal{N}_{I}$ denotes the restoration network, $\mathcal{C}$ denotes the concatenation operation, and $\mathcal{S}$ denotes the space-to-depth transformation.
We use the following loss function to constrain the network $\mathcal{N}_{I}$:
\begin{equation}
\vspace{-1mm}
\mathcal{L} = \|I_i^* - I_i\|_1.
\label{eq: loss-function-whole}
\vspace{-1mm}
\end{equation}
%

\vspace{-1mm}
\subsection{Implementation details}
\label{ssec: Parameter settings and training data}
\vspace{-1mm}
{\flushleft \bf{Training datasets.}}
We train the proposed algorithm using the REDS dataset~\cite{REDS}, where the REDS dataset contains 300 videos, each video contains 100 frames with an image size of $720\times 1280$ pixels.
Among 300 videos, 240 videos are used for training, 30 videos are used for validation, and the remaining 30 videos are used for test.
During the training, we randomly choose 45 consecutive frames from each video in the training dataset to train the proposed algorithm.
\vspace{-5mm}
{\flushleft \bf{Parameter settings and training details.}}
We empirically set $\gamma = 0.02$.
We use the similar data augmentation method to~\cite{edvr} to generate training data. The batch size is set to be $8$, and the size of each image patch is $64\times 64$ pixels.
In the training process, we use the ADAM optimizer~\cite{adam} with parameters $\beta_1 = 0.9$, $\beta_2 = 0.999$, and $\epsilon = 10^{-8}$.
The motion blur estimation network $\mathcal{N}_k$ takes the Gaussian kernels as the input, where the settings of the Gaussian kernels are the same as~\cite{shan/sr/sa08}.
The size of Gaussian kernel is empirically set to be $15\times 15$ pixels.
The sizes of the two fully connected layers are set to be $1000$ and $225$, respectively. The output size of $\mathcal{N}_k$ is set to be $15\times 15$ pixels.
The optical flow estimation network $\mathcal{N}_o$ is initialized by the pre-trained model~\cite{pwcnent/deqing}.
Both the kernel estimation network $\mathcal{N}_k$ and the image restoration network $\mathcal{N}_I$ use the random initialization and are trained from scratch.
The learning rates for both kernel estimation network $\mathcal{N}_k$ and image restoration network $\mathcal{N}_I$ are initialized to be $10^{-4}$.
As we use the pre-trained model~\cite{pwcnent/deqing} to initialize the optical flow estimation network, the learning rate for this network is initialized to be $10^{-6}$.
All the learning rates decrease to ${0.2}$ times after every 50 epochs.
During the training process, we first train $\mathcal{N}_k$ and then jointly train $\mathcal{N}_o$ and $\mathcal{N}_I$ in an end-to-end manner.
The algorithm is implemented based on the PyTorch. More experimental results are included in the supplemental material.
The training code and test model are available at~\url{https://github.com/jspan/blindvsr}.

\begin{table*}[!t]
\vspace{-1mm}
  \caption{Quantitative evaluations on the REDS dataset~\cite{REDS} in terms of PSNR and SSIM. All the results are generated according to the published models for fair comparisons. The best two results are shown in \textcolor[rgb]{1.00,0.00,0.00}{\textbf{red}} and \textcolor[rgb]{0.00,0.00,1.00}{\underline{blue}}.
  }
   \vspace{1mm}
   \label{tab: results-reds}
\footnotesize
 \centering
 \begin{tabular}{lcccccccccccc}
    \toprule
    Methods                &Bicubic  &RCAN~\cite{RCAN}  & SPMC~\cite{xintao/iccv17}  &DUF~\cite{jo/cvpr18}   &TOF~\cite{tof} &RBPN~\cite{VDBPN/cvpr19}  &EDVR-M~\cite{edvr} &EDVR~\cite{edvr}                &Ours  \\
    \hline
     Avg. PSNRs        &25.59        &28.71     &27.74 &28.60                &27.77          &29.82           & 30.51     &\textcolor[rgb]{1.00,0.00,0.00}{\textbf{31.07 }}               &\textcolor[rgb]{0.00,0.00,1.00}{\underline{30.51}}         \\
     Avg. SSIMs        &0.7077       &0.8184    &0.7915  &0.8254               &0.7949         &0.8537         &0.8699     & \textcolor[rgb]{1.00,0.00,0.00}{\textbf{0.8802}}              &\textcolor[rgb]{0.00,0.00,1.00}{\underline{0.8674}}        \\

 \bottomrule
  \end{tabular}
\vspace{-3mm}
\end{table*}

\begin{figure*}[!t]\footnotesize
\begin{center}
\begin{tabular}{cccccccc}
\multicolumn{3}{c}{\multirow{5}*[74pt]{\includegraphics[width=0.368\linewidth, height = 0.352\linewidth]{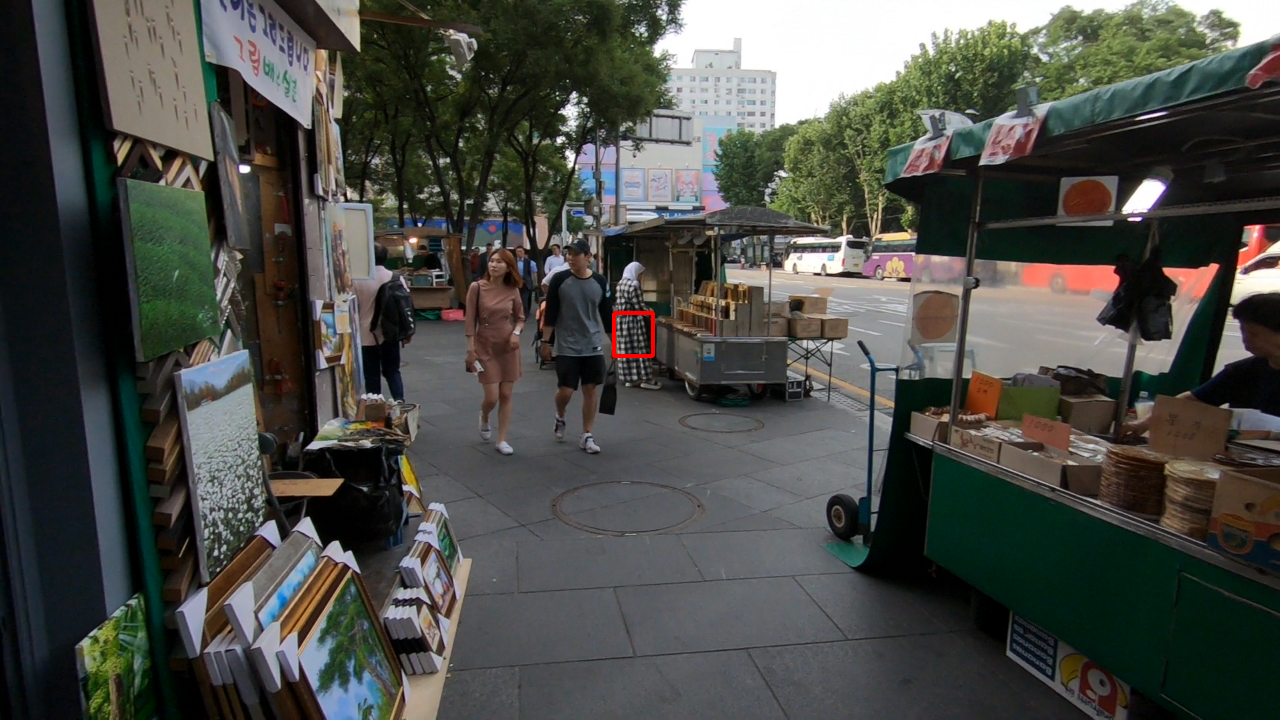}}}&\hspace{-4.5mm}
\includegraphics[width=0.15\linewidth]{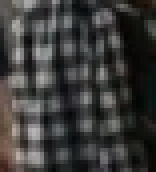} &\hspace{-4.5mm}
\includegraphics[width=0.15\linewidth]{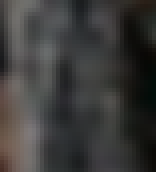} &\hspace{-4.5mm}
\includegraphics[width=0.15\linewidth]{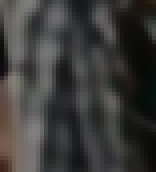} &\hspace{-4.5mm}
\includegraphics[width=0.15\linewidth]{figures/reds20/00000010_rcan_crop_x4} \\
\multicolumn{3}{c}{~} &\hspace{-4.5mm}  (b) HR patch &\hspace{-4.5mm}  (c) Bicubic &\hspace{-4.5mm}  (d) RCAN~\cite{RCAN}  &\hspace{-4.5mm}  (e) SPMC~\cite{xintao/iccv17}\\
%
\multicolumn{3}{c}{~} & \hspace{-4.5mm}
\includegraphics[width=0.15\linewidth]{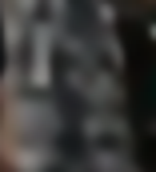} & \hspace{-4.5mm}
\includegraphics[width=0.15\linewidth]{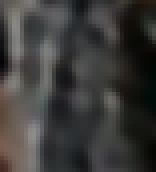} & \hspace{-4.5mm}
\includegraphics[width=0.15\linewidth]{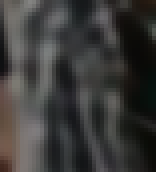} & \hspace{-4.5mm}
\includegraphics[width=0.15\linewidth]{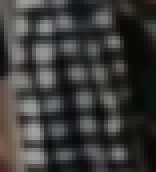} \\
\multicolumn{3}{c}{\hspace{-4.5mm} (a) Ground truth HR image} &  \hspace{-4.5mm} (f) DUF~\cite{jo/cvpr18} &\hspace{-4.5mm}  (g) TOFlow~\cite{tof}&\hspace{-4.5mm}  (h) RBPN~\cite{VDBPN/cvpr19}  & \hspace{-4.5mm} (i) Ours\\
\end{tabular}
\end{center}
\vspace{-3mm}
\caption{Video SR result ($\times 4$) on the REDS dataset~\cite{REDS}. The proposed algorithm recovers high-quality images with clearer structures.
}%
\label{fig: results-reds}
\vspace{-0mm}
\end{figure*}

\begin{table*}[!t]
\vspace{-1mm}
  \caption{Quantitative evaluations on the Vid4 dataset~\cite{Bayesian/vsr/tpami14} and SPMCS dataset~\cite{xintao/iccv17} in terms of PSNR and SSIM.
  All the results are generated according to the published models for fair comparisons. * means the values from the reported results~\cite{tof}. The best two results are shown in \textcolor[rgb]{1.00,0.00,0.00}{\textbf{red}} and \textcolor[rgb]{0.00,0.00,1.00}{\underline{blue}}.
  }
   \vspace{1mm}
   \label{tab: result-datasets-vid4-spmc}
\footnotesize
 \centering
 \begin{tabular}{l|ccccccccccccc}
    \toprule
& Methods                                                   &Bicubic   &  BayesianSR~\cite{maziyang/vsr/cvpr15}* &RCAN~\cite{RCAN} & SPMC~\cite{xintao/iccv17} &DUF~\cite{jo/cvpr18} &TOFlow~\cite{tof}  &Ours\\
\hline
\multirow{2}{*}{Vid4 dataset~\cite{Bayesian/vsr/tpami14}}  &Avg. PSNRs        &21.91    & 21.95     &24.03      &24.39        &\textcolor[rgb]{1.00,0.00,0.00}{\bf{25.84}}        &24.22                               &\textcolor[rgb]{0.00,0.00,1.00}{\underline{25.35}}         \\
                                                           &Avg. SSIMs        &0.5825   &0.7369    &0.7206      &0.7534       &\textcolor[rgb]{1.00,0.00,0.00}{\bf{0.8151}}         &0.7396                          &\textcolor[rgb]{0.00,0.00,1.00}{\underline{0.7868}}        \\
                                                             \hline
                                                             \hline
     \multirow{2}{*}{SPMCS dataset~\cite{xintao/iccv17}}         &Avg. PSNRs        &25.16    &-    &28.60       &28.19       &\textcolor[rgb]{0.00,0.00,1.00}{\underline{29.31}}        &27.62                     &\textcolor[rgb]{1.00,0.00,0.00}{\bf{29.54}}         \\
                                                                &Avg. SSIMs        &0.6962   &-    &0.8253      &0.8164      &\textcolor[rgb]{1.00,0.00,0.00}{\textbf{0.8554}}       &0.8048                      &\textcolor[rgb]{0.00,0.00,1.00}{\underline{0.8532}}        \\
 \bottomrule
  \end{tabular}
\vspace{-5mm}
\end{table*}
%

\begin{figure*}[!t]\footnotesize
\begin{center}
\begin{tabular}{cccccccc}
\multicolumn{3}{c}{\multirow{5}*[52pt]{\includegraphics[width=0.368\linewidth, height = 0.263\linewidth]{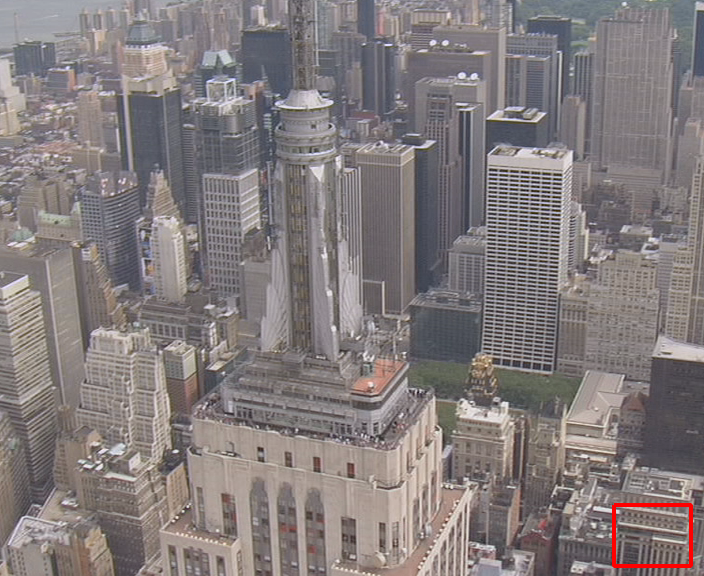}}}&\hspace{-4.5mm}
\includegraphics[width=0.15\linewidth]{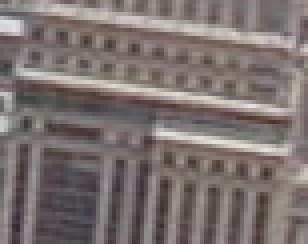} &\hspace{-4.5mm}
\includegraphics[width=0.15\linewidth]{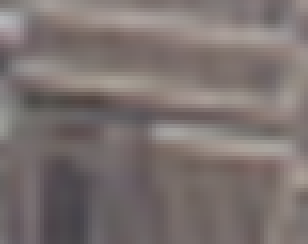} &\hspace{-4.5mm}
\includegraphics[width=0.15\linewidth]{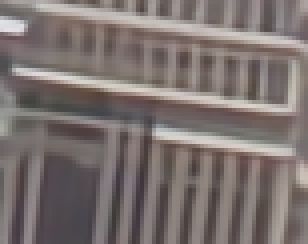} &\hspace{-4.5mm}
\includegraphics[width=0.15\linewidth]{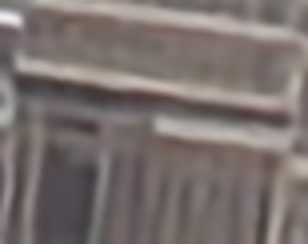} \\
\multicolumn{3}{c}{~} &\hspace{-4.5mm}  (b1) HR patch &\hspace{-4.5mm}  (c1) Bicubic &\hspace{-4.5mm}  (d1) RCAN~\cite{RCAN}  &\hspace{-4.5mm}  (e1) SPMC~\cite{xintao/iccv17}\\
\multicolumn{3}{c}{~} & \hspace{-4.5mm}
\includegraphics[width=0.15\linewidth]{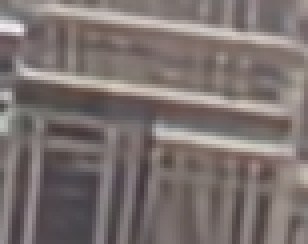} & \hspace{-4.5mm}
\includegraphics[width=0.15\linewidth]{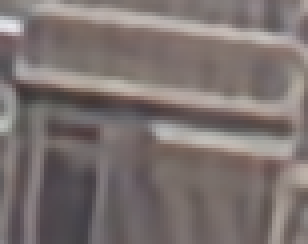} & \hspace{-4.5mm}
\includegraphics[width=0.15\linewidth]{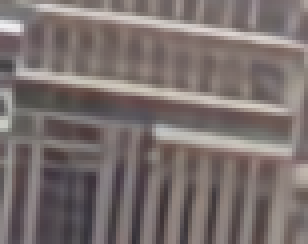} & \hspace{-4.5mm}
\includegraphics[width=0.15\linewidth]{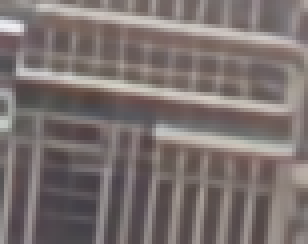} \\
\multicolumn{3}{c}{\hspace{-4.5mm} (a1) Ground truth HR image} &  \hspace{-4.5mm} (f1) DUF~\cite{jo/cvpr18}  &\hspace{-4.5mm}  (g1) TOFlow~\cite{tof} &\hspace{-4.5mm}  (h1) RBPN~\cite{VDBPN/cvpr19} & \hspace{-4.5mm} (i1) Ours\\
\multicolumn{3}{c}{\multirow{5}*[64pt]{\includegraphics[width=0.368\linewidth, height = 0.31\linewidth]{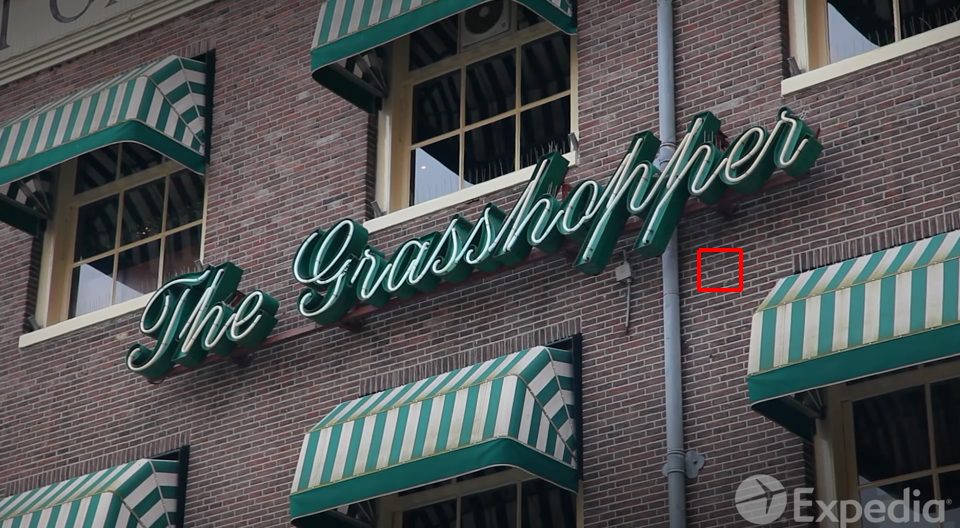}}}&\hspace{-4.5mm}
\includegraphics[width=0.15\linewidth]{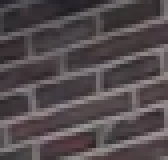} &\hspace{-4.5mm}
\includegraphics[width=0.15\linewidth]{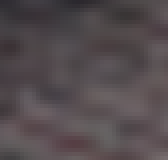} &\hspace{-4.5mm}
\includegraphics[width=0.15\linewidth]{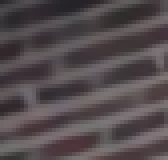} &\hspace{-4.5mm}
\includegraphics[width=0.15\linewidth]{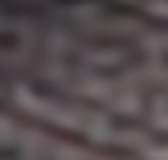} \\
\multicolumn{3}{c}{~} &\hspace{-4.5mm}  (b1) HR patch &\hspace{-4.5mm}  (c2) Bicubic &\hspace{-4.5mm}  (d2) RCAN~\cite{RCAN}  &\hspace{-4.5mm}  (e2) SPMC~\cite{xintao/iccv17}\\
\multicolumn{3}{c}{~} & \hspace{-4.5mm}
\includegraphics[width=0.15\linewidth]{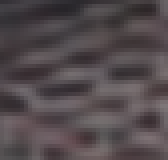} & \hspace{-4.5mm}
\includegraphics[width=0.15\linewidth]{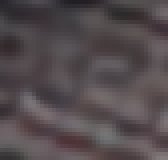} & \hspace{-4.5mm}
\includegraphics[width=0.15\linewidth]{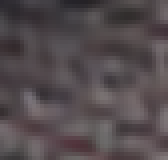} & \hspace{-4.5mm}
\includegraphics[width=0.15\linewidth]{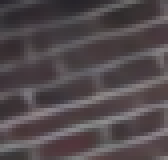} \\
\multicolumn{3}{c}{\hspace{-4.5mm} (a2) Ground truth HR image} &  \hspace{-4.5mm} (f2) DUF~\cite{jo/cvpr18}  &\hspace{-4.5mm}  (g2) TOFlow~\cite{tof} &\hspace{-4.5mm}  (h2) RBPN~\cite{VDBPN/cvpr19} & \hspace{-4.5mm} (i2) Ours\\
\end{tabular}
\end{center}
\vspace{-2mm}
\caption{Video SR results ($\times 4$) on the Vid4~\cite{Bayesian/vsr/tpami14} and SPMCS~\cite{xintao/iccv17} datasets. The proposed algorithm generates much clearer images.
}%
\label{fig: results-vid4-spmc}
\vspace{-5mm}
\end{figure*}

\section{Experimental Results}
In this section, we compare the proposed algorithm against state-of-the-art methods using publicly available benchmark datasets.
\vspace{-2mm}
{\flushleft \bf{Quantitative evaluations.}}
We compare the proposed algorithm against state-of-the-art methods including the variational methods~\cite{maziyang/vsr/cvpr15}
and deep CNN-based methods including DUF~\cite{jo/cvpr18}, TOFlow~\cite{tof}, RBPN~\cite{VDBPN/cvpr19}, EDVR~\cite{edvr}.
In addition, we compare the proposed method with state-of-the-art deep CNNs-based single image SR~\cite{RCAN} (RCAN).
We use the PSNR and SSIM as the evaluation metrics to evaluate the quality of each restored image on synthetic datasets.
The PSNR and SSIM values of each restored image are calculated using RGB channels based on the script by~\cite{edvr}.

Table~\ref{tab: results-reds} shows the quantitative evaluation results on 4 videos from the REDS test dataset~\cite{REDS}, where these 4 videos are also used in~\cite{edvr} for test.
Overall, the proposed method achieves comparable results compared to the EDVR algorithm and outperforms other algorithms by a large margin.
Figure~\ref{fig: results-reds} shows some results with a scale factor of $4$ by the top-performing methods on the REDS dataset~\cite{REDS}.
We note that the state-of-the-art single image SR method~\cite{RCAN} does not recover the structural details well as shown in Figure~\ref{fig: results-reds}(d).
%
%
Tao et al.~\cite{xintao/iccv17} develop an effective warping layer for video SR. However, this structural details in the super-resolved image are not sharp.
Xue et al.~\cite{tof} explicitly use optical flow and warping operations for video restoration.
However, this algorithm assumes the blur kernel is known and takes the Bicubic upsampled images as inputs which are blurry (Figure~\ref{fig: intermediate-result-kernel}(b)) and thus accordingly affect the details restoration (Figure~\ref{fig: results-reds}(g)).
Jo et al.~\cite{jo/cvpr18} develop an effective algorithm to dynamically estimate upsampling filters and residual images for video SR. However, the structural details are not restored well due to the inaccurate upsampling filters (Figure~\ref{fig: results-reds}(h)).
%
%
As the proposed algorithm develops a motion blur estimation which provides an intermediate latent HR image with sharp contents, it generates much clearer images with finer details (Figure~\ref{fig: results-reds}(i)).

We then evaluate the proposed algorithm on the test dataset by Liu et al.~\cite{Bayesian/vsr/tpami14} (Vid4) and Tao et al.~\cite{xintao/iccv17} (SPMCS).
Table~\ref{tab: result-datasets-vid4-spmc} shows the quantitative results on the Vid4 and SPMCS datasets.
We note that the variational model-based method~\cite{Bayesian/vsr/tpami14} estimates blur kernels from video sequences to restore images. However, the performance of these methods is limited by the hand-crafted image priors.
The deep CNN-based methods~\cite{xintao/iccv17,jo/cvpr18,tof,VDBPN/cvpr19} generate the results with higher PSNR and SSIM values than the variational model-based methods.
In contrast, our algorithm generates favorable results in terms of PSNR and SSIM due to the use of motion blur estimation.
%

Figure~\ref{fig: results-vid4-spmc} shows some SR results with a scale factor of 4 by the top-performing methods on the Vid4~\cite{Bayesian/vsr/tpami14} and SPMCS~\cite{xintao/iccv17} datasets.
State-of-the-art methods do not recover the structural details well.
In contrast, the proposed method jointly estimates motion blur, motion fields, and latent images. The motion blur estimation is able to generate intermediate HR image with clear details, which thus lead to much clearer images with finer structural details.

%

\begin{figure*}[!t]\footnotesize
\begin{center}
\begin{tabular}{cccccccc}
\multicolumn{3}{c}{\multirow{5}*[64pt]{\includegraphics[width=0.368\linewidth, height = 0.31\linewidth]{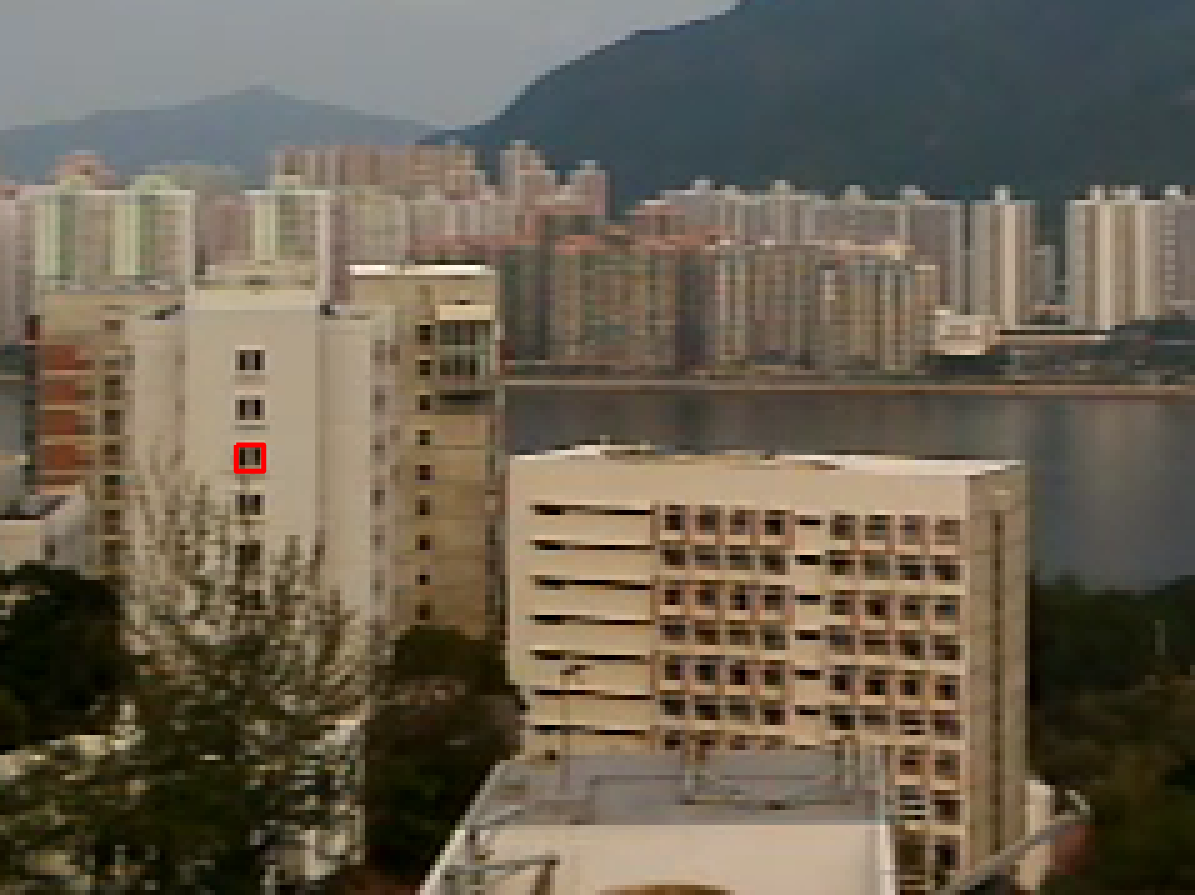}}}&\hspace{-4.5mm}
\includegraphics[width=0.15\linewidth]{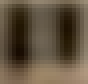} &\hspace{-4.5mm}
\includegraphics[width=0.15\linewidth]{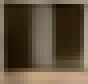} &\hspace{-4.5mm}
\includegraphics[width=0.15\linewidth]{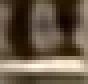} &\hspace{-4.5mm}
\includegraphics[width=0.15\linewidth]{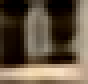} \\
\multicolumn{3}{c}{~} &\hspace{-4.5mm}  (b1) Bicubic &\hspace{-4.5mm}  (c1) RCAN~\cite{RCAN} &\hspace{-4.5mm}  (d1)  SPMC~\cite{xintao/iccv17} &\hspace{-4.5mm}  (e1) DUF~\cite{jo/cvpr18} \\
\multicolumn{3}{c}{~} & \hspace{-4.5mm}
\includegraphics[width=0.15\linewidth]{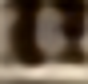} & \hspace{-4.5mm}
\includegraphics[width=0.15\linewidth]{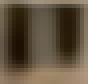} & \hspace{-4.5mm}
\includegraphics[width=0.15\linewidth]{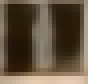} & \hspace{-4.5mm}
\includegraphics[width=0.15\linewidth]{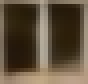} \\
\multicolumn{3}{c}{\hspace{-4.5mm} (a1) Input} &  \hspace{-4.5mm} (f1) TOF~\cite{tof} &\hspace{-4.5mm}  (g1) RBPN~\cite{VDBPN/cvpr19} &\hspace{-4.5mm}  (h1) EDVR~\cite{edvr} & \hspace{-4.5mm} (i1) Ours\\
\multicolumn{3}{c}{\multirow{5}*[59pt]{\includegraphics[width=0.368\linewidth, height = 0.288\linewidth]{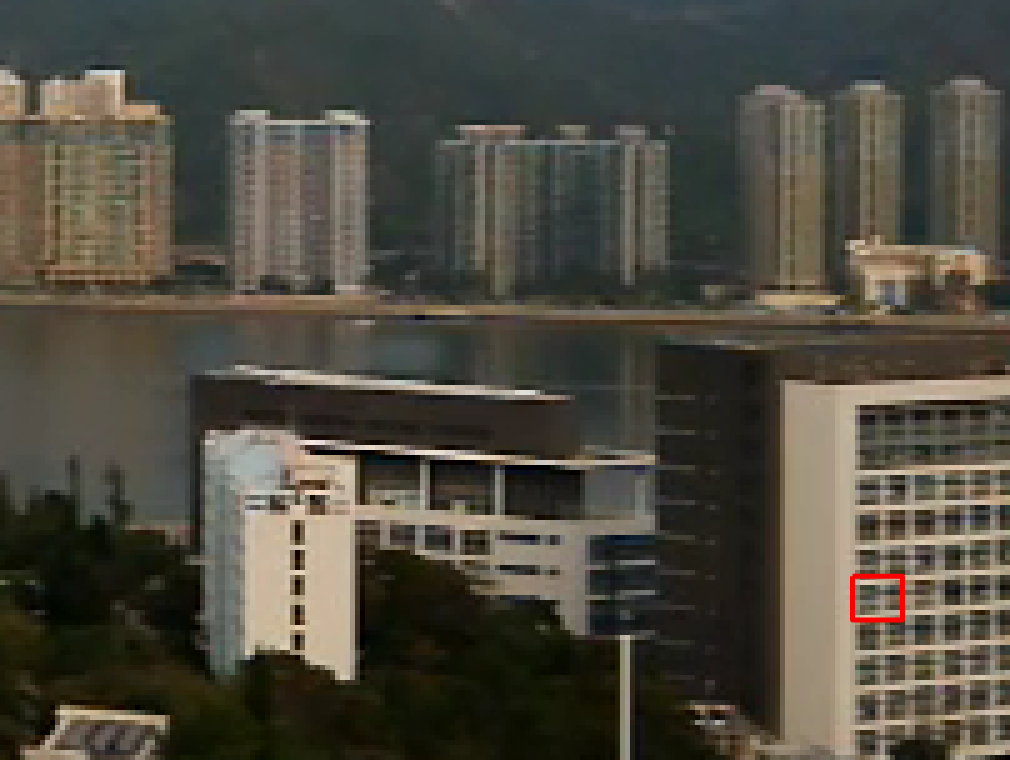}}}&\hspace{-4.5mm}
\includegraphics[width=0.15\linewidth]{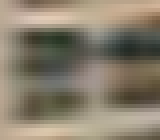} &\hspace{-4.5mm}
\includegraphics[width=0.15\linewidth]{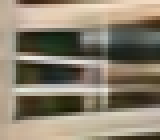} &\hspace{-4.5mm}
\includegraphics[width=0.15\linewidth]{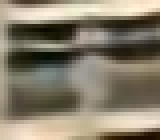} &\hspace{-4.5mm}
\includegraphics[width=0.15\linewidth]{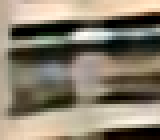} \\
\multicolumn{3}{c}{~} &\hspace{-4.5mm}  (b2) Bicubic &\hspace{-4.5mm}  (c2) RCAN~\cite{RCAN} &\hspace{-4.5mm}  (d2)  SPMC~\cite{xintao/iccv17} &\hspace{-4.5mm}  (e2) DUF~\cite{jo/cvpr18} \\
\multicolumn{3}{c}{~} & \hspace{-4.5mm}
\includegraphics[width=0.15\linewidth]{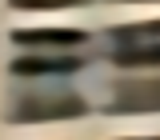} & \hspace{-4.5mm}
\includegraphics[width=0.15\linewidth]{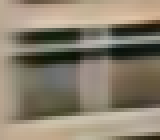} & \hspace{-4.5mm}
\includegraphics[width=0.15\linewidth]{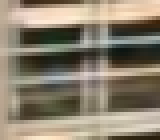} & \hspace{-4.5mm}
\includegraphics[width=0.15\linewidth]{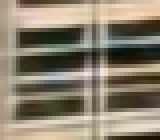} \\
\multicolumn{3}{c}{\hspace{-4.5mm} (a2) Input} &  \hspace{-4.5mm} (f2) TOFlow~\cite{tof} &\hspace{-4.5mm}  (g2) RBPN~\cite{VDBPN/cvpr19} &\hspace{-4.5mm}  (h2) EDVR~\cite{edvr} & \hspace{-4.5mm} (i2) Ours\\
\end{tabular}
\end{center}
\vspace{-3mm}
\caption{Video SR results on real videos ($\times 4$). The proposed algorithm generates much clearer images.
}%
\label{fig: results-real-videos}
\vspace{-3mm}
\end{figure*}

\vspace{-2mm}
{\flushleft \bf{Qualitative evaluations.}}
We further qualitatively evaluate the proposed algorithm against state-of-the-art methods on real videos.
Figure~\ref{fig: results-real-videos} shows a real example from~\cite{liao/iccv15}.
The results by state-of-the-art methods~\cite{jo/cvpr18,VDBPN/cvpr19,tof,edvr} are still blurry.
In contrast, our algorithm generates the images with clearer detailed structures, which demonstrate that the proposed algorithm generalizes well.

\vspace{-1mm}
\section{Analysis and Discussions}
\label{sec: Analysis and Discussions}
We have shown that using the motion blur estimation is able to help details restoration in video SR.
In this section, we further analyze the effect of the proposed algorithm.
%
\begin{table}[!t]
  \caption{Effectiveness of the motion blur estimation on video SR ($\times 4$). The results are obtained from the REDS dataset~\cite{REDS}.
  }
   \label{tab: baselines-kernel}
\footnotesize
 \centering
 \begin{tabular}{lcccccccccc}
    \toprule
   Methods                                                 &BaselineLR                 &BaselineHR             &Ours      \\
    \hline
 Avg. PSNRs                                          &30.18         &30.48            &\textcolor[rgb]{1.00,0.00,0.00}{\textbf{30.51}}      \\
 Avg. SSIMs                                          &0.8609         &0.8669          &\textcolor[rgb]{1.00,0.00,0.00}{\textbf{0.8674}}     \\
 \bottomrule
  \end{tabular}
\vspace{-4.0mm}
\end{table}
%

\begin{figure}[!t]\footnotesize
\centering
\begin{tabular}{ccc}
\includegraphics[width=0.32\linewidth]{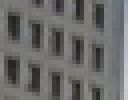} &\hspace{-4mm}
\includegraphics[width=0.32\linewidth]{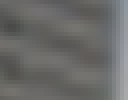} &\hspace{-4mm}
\includegraphics[width=0.32\linewidth]{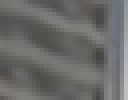}  \\
(a) HR patch &\hspace{-4mm}  (b) Bicubic & \hspace{-4mm} (c)  DUF~\cite{jo/cvpr18} \\
\includegraphics[width=0.32\linewidth]{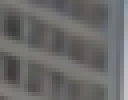} &\hspace{-4mm}
\includegraphics[width=0.32\linewidth]{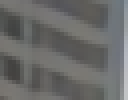} &\hspace{-4mm}
\includegraphics[width=0.32\linewidth]{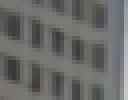}  \\
(d) BaselineLR &\hspace{-4mm}  (e) BaselineHR &\hspace{-4mm}  (f) Ours \\
\end{tabular}
\caption{Effectiveness of the motion blur estimation on video SR ($\times$4). Using motion blur estimation is able to generate the results with clearer structural details.}%
\label{fig: baseline-analysis}
\vspace{-4mm}
\end{figure}

%
\vspace{-3mm}
{\flushleft \bf{Effectiveness of the motion blur estimation.}}
The proposed motion blur estimation process provides blur kernels which thus leads to the intermediate latent images with clear contents for better details restoration.
To demonstrate the effectiveness of this method, we disable this step in the proposed algorithm for fair comparisons.
Thus, the proposed method reduces to the method without using the motion blur estimation and intermediate latent image restoration.
For this case, the inputs of this baseline method can be either the bicubic upsampling results (i.e., $I_i^{b}$, $I_{i+1}^{b}$, and $I_{i-1}^b$) (BaselineHR for short) or the original LR images (BaselineLR for short).

Table~\ref{tab: baselines-kernel} shows the quantitative evaluations on the REDS dataset~\cite{REDS}.
The average PSNR by our method is 0.33dB higher than that by BaselineLR, which demonstrates that using motion blur estimation is able to generate much better results.

The visualizations in Figure~\ref{fig: baseline-analysis} further demonstrate that directly estimating the HR images using deep CNN models without motion blur estimation does not generates the results with clearer structural details, while the proposed method generates much clearer images.

\vspace{-2mm}
\begin{table}[!t]
  \caption{Accuracy of the estimated motion blur kernels.
  }
   \label{tab: kernel-accuracy}
\footnotesize
 \centering
 \begin{tabular}{lcccccccccc}
    \toprule
   Methods                                                 &Ma et al.~\cite{maziyang/vsr/cvpr15}        &Gaussian kernel            &Ours\\
   \hline
   Avg. PSNRs                                              &31.24        &25.48      &\textcolor[rgb]{1.00,0.00,0.00}{\textbf{42.06}}              \\
   Avg. SSIMs                                              &0.9272       &0.8726     &\textcolor[rgb]{1.00,0.00,0.00}{\textbf{0.9962}}          \\
 \bottomrule
  \end{tabular}
\vspace{-5mm}
\end{table}

%
\vspace{-2mm}
{\flushleft \bf{Analysis on motion blur kernels.}}
%
Different from existing video SR methods that use known blur kernels (e.g., Gaussian blur kernel~\cite{gaussian/kernel/cvpr18}) and do not model them in the video SR process,
we develop a motion blur estimation method to estimate blur kernels for video SR.
To examine accuracy of the estimated blur kernels, we apply the estimated blur kernels and downsampling operation to the ground truth HR images to generate LR images.
We apply the Bicubic downsampling to the HR images to obtain the LR ones as the ground truth LR images.
The quality of the regenerated LR images is used to measure the accuracy of the estimated blur kernels.
Table~\ref{tab: kernel-accuracy} shows that regenerated LR images by the proposed method are closed to the ground truth LR images, which indicates the proposed algorithm is able to estimate more accurate blur kernels than those conventional variational model-based method~\cite{maziyang/vsr/cvpr15}.

\begin{figure}[!t]\footnotesize
\centering
\begin{tabular}{ccc}
\hspace{-2mm}
\includegraphics[width=0.32\linewidth]{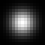} &
\includegraphics[width=0.32\linewidth]{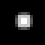} \\
(a) Initial kernel  &  (b) Estimated kernel\\
\end{tabular}
\caption{Visualizations of the estimated motion blur kernels generated by the network $\mathcal{N}_k$. We use the Bicubic downampling as the degradation process of video SR for test.}
\label{fig: intermediate-kernels}
\vspace{-4mm}
\end{figure}

Figure~\ref{fig: intermediate-kernels} shows the visualizations of estimated blur kernels by $\mathcal{N}_k$ when the degradation process is the Bicubic downampling.
We note that shape of the estimated blur kernels are similar to that of Bicubic kernel as demonstrated in~\cite{levin/etal/blindsr}. Thus, both the quantitative and qualitative results demonstrate that the proposed algorithm is able to capture the degradation process well.

%
\vspace{-3mm}
{\flushleft \bf{Video SR with motion blur.}}
The proposed algorithm is able to super-resolve videos containing small motion blur to some extent.
Figure~\ref{fig: results-real-videos}(a) shows a LR frame from a LR video in~\cite{maziyang/vsr/cvpr15}, where the LR video contains small motion blur.
As most video SR methods~\cite{xintao/iccv17,jo/cvpr18,tof,VDBPN/cvpr19,edvr} do not model the motion blur, the generated results are blurry.
Although the method~\cite{maziyang/vsr/cvpr15} is able to solve video SR with motion blur, it is limited to the hand-crafted priors.
In contrast, the proposed method generates much clearer images.

\section{Concluding Remarks}
   We have proposed an effective video super-resolution algorithm. The proposed algorithm consists of motion blur estimation, motion field estimation, and latent image restoration modules.
   The motion blur estimation module is able to provide reliable blur kernels.
   With the estimated blur kernel, we develop an image deconvolution method based on the image formation model of video super-resolution to generate intermediate latent images
   so that some sharp image contents can be restored well.
   To generate high-quality images, we use the motion estimation module to explore the information from adjacent frames to constrain the deep CNN model for better image restoration.
   We have shown that the proposed method is able to generate much clearer images with finer structural details due to the use of motion blur kernel estimation. Both quantitative evaluations and qualitative evaluations show that the proposed algorithm performs favorably against state-of-the-art methods.

\clearpage
{\small
\bibliographystyle{ieee_fullname}
\bibliography{egbib}

\begin{thebibliography}{10}\itemsep=-1pt

\bibitem{prior/eccv96}
Benedicte Bascle, Andrew Blake, and Andrew Zisserman.
\newblock Motion deblurring and super-resolution from an image sequence.
\newblock In {\em ECCV}, pages 573--582, 1996.

\bibitem{GAN/data/eccv18}
Adrian Bulat, Jing Yang, and Georgios Tzimiropoulos.
\newblock To learn image super-resolution, use a {GAN} to learn how to do image
  degradation first.
\newblock In {\em ECCV}, pages 187--202, 2018.

\bibitem{vespcn}
Jose Caballero, Christian Ledig, Andrew~P. Aitken, Alejandro Acosta, Johannes
  Totz, Zehan Wang, and Wenzhe Shi.
\newblock Real-time video super-resolution with spatio-temporal networks and
  motion compensation.
\newblock In {\em CVPR}, pages 2848--2857, 2017.

\bibitem{prior/tip01}
Stephen~C. Cain, Majeed~M. Hayat, and Ernest~E. Armstrong.
\newblock Projection-based image registration in the presence of fixed-pattern
  noise.
\newblock {\em {IEEE} TIP}, 10(12):1860--1872, 2001.

\bibitem{TecoGAN}
Mengyu Chu, You Xie, Laura Leal-Taix{\'{e}}, and Nils Thuerey.
\newblock Temporally coherent gans for video super-resolution (tecogan).
\newblock {\em CoRR}, abs/1811.09393, 2018.

\bibitem{SRCNN/tpmai}
Chao Dong, Chen~Change Loy, Kaiming He, and Xiaoou Tang.
\newblock Image super-resolution using deep convolutional networks.
\newblock {\em {IEEE} TPAMI}, 38(2):295--307, 2016.

\bibitem{Accelerating/dong}
Chao Dong, Chen~Change Loy, and Xiaoou Tang.
\newblock Accelerating the super-resolution convolutional neural network.
\newblock In {\em ECCV}, pages 391--407, 2016.

\bibitem{levin/etal/blindsr}
Netalee Efrat, Daniel Glasner, Alexander Apartsin, Boaz Nadler, and Anat Levin.
\newblock Accurate blur models vs. image priors in single image
  super-resolution.
\newblock In {\em ICCV}, pages 2832--2839, 2013.

\bibitem{Milanfar/tip04}
Sina Farsiu, M.~Dirk Robinson, Michael Elad, and Peyman Milanfar.
\newblock Fast and robust multiframe super resolution.
\newblock {\em {IEEE} TIP}, 13(10):1327--1344, 2004.

\bibitem{Gu/cvpr19}
Jinjin Gu, Hannan Lu, Wangmeng Zuo, and Chao Dong.
\newblock Blind super-resolution with iterative kernel correction.
\newblock In {\em CVPR}, pages 1604--1613, 2019.

\bibitem{DBPN}
Muhammad Haris, Gregory Shakhnarovich, and Norimichi Ukita.
\newblock Deep back-projection networks for super-resolution.
\newblock In {\em CVPR}, pages 1664--1673, 2018.

\bibitem{VDBPN/cvpr19}
Muhammad Haris, Gregory Shakhnarovich, and Norimichi Ukita.
\newblock Recurrent back-projection network for video super-resolution.
\newblock In {\em CVPR}, pages 3897--3906, 2019.

\bibitem{Huang/nips15}
Yan Huang, Wei Wang, and Liang Wang.
\newblock Bidirectional recurrent convolutional networks for multi-frame
  super-resolution.
\newblock In {\em NeurIPS}, pages 235--243, 2015.

\bibitem{jo/cvpr18}
Younghyun Jo, Seoung~Wug Oh, Jaeyeon Kang, and Seon~Joo Kim.
\newblock Deep video super-resolution network using dynamic upsampling filters
  without explicit motion compensation.
\newblock In {\em CVPR}, pages 3224--3232, 2018.

\bibitem{Kappeler/tci16}
Armin Kappeler, Seunghwan Yoo, Qiqin Dai, and Aggelos~K. Katsaggelos.
\newblock Video super-resolution with convolutional neural networks.
\newblock {\em {IEEE} TCI}, 2(2):109--122, 2016.

\bibitem{VDSR}
Jiwon Kim, Jung~Kwon Lee, and Kyoung~Mu Lee.
\newblock Accurate image super-resolution using very deep convolutional
  networks.
\newblock In {\em CVPR}, pages 1646--1654, 2016.

\bibitem{adam}
Diederik~P. Kingma and Jimmy Ba.
\newblock Adam: {A} method for stochastic optimization.
\newblock {\em CoRR}, abs/1412.6980, 2014.

\bibitem{SRGAN}
Christian Ledig, Lucas Theis, Ferenc Huszar, Jose Caballero, Andrew Cunningham,
  Alejandro Acosta, Andrew~P. Aitken, Alykhan Tejani, Johannes Totz, Zehan
  Wang, and Wenzhe Shi.
\newblock Photo-realistic single image super-resolution using a generative
  adversarial network.
\newblock In {\em CVPR}, pages 105--114, 2017.

\bibitem{liao/iccv15}
Renjie Liao, Xin Tao, Ruiyu Li, Ziyang Ma, and Jiaya Jia.
\newblock Video super-resolution via deep draft-ensemble learning.
\newblock In {\em ICCV}, pages 531--539, 2015.

\bibitem{edsr}
Bee Lim, Sanghyun Son, Heewon Kim, Seungjun Nah, and Kyoung~Mu Lee.
\newblock Enhanced deep residual networks for single image super-resolution.
\newblock In {\em {CVPR}}, pages 1132--1140, 2017.

\bibitem{Bayesian/vsr/tpami14}
Ce Liu and Deqing Sun.
\newblock On bayesian adaptive video super resolution.
\newblock {\em {IEEE} TPAMI}, 36(2):346--360, 2014.

\bibitem{liu/iccv17}
Ding Liu, Zhaowen Wang, Yuchen Fan, Xianming Liu, Zhangyang Wang, Shiyu Chang,
  and Thomas~S. Huang.
\newblock Robust video super-resolution with learned temporal dynamics.
\newblock In {\em ICCV}, pages 2526--2534, 2017.

\bibitem{GAN/video/tip19}
Alice Lucas, Santiago~Lopez Tapia, Rafael Molina, and Aggelos~K. Katsaggelos.
\newblock Generative adversarial networks and perceptual losses for video
  super-resolution.
\newblock {\em {IEEE} TIP}, 28(7):3312--3327, 2019.

\bibitem{maziyang/vsr/cvpr15}
Ziyang Ma, Renjie Liao, Xin Tao, Li Xu, Jiaya Jia, and Enhua Wu.
\newblock Handling motion blur in multi-frame super-resolution.
\newblock In {\em CVPR}, pages 5224--5232, 2015.

\bibitem{Tomer/blindsr/iccv13}
Tomer Michaeli and Michal Irani.
\newblock Nonparametric blind super-resolution.
\newblock In {\em ICCV}, pages 945--952, 2013.

\bibitem{REDS}
Seungjun Nah, Sungyong Baik, Seokil Hong, Gyeongsik Moon, Sanghyun Son, Radu
  Timofte, and Kyoung~Mu Lee.
\newblock {NTIRE} 2019 challenge on video deblurring and super-resolution:
  Dataset and study.
\newblock In {\em {CVPR} Workshops}, 2019.

\bibitem{dongwei/deepdeblur}
Dongwei Ren, Kai Zhang, Qilong Wang, Qinghua Hu, and Wangmeng Zuo.
\newblock Neural blind deconvolution using deep priors.
\newblock {\em CoRR}, abs/1908.02197, 2019.

\bibitem{Sajjadi_2017_ICCV}
Mehdi S.~M. Sajjadi, Bernhard Scholkopf, and Michael Hirsch.
\newblock Enhancenet: Single image super-resolution through automated texture
  synthesis.
\newblock In {\em ICCV}, pages 4491--4500, 2017.

\bibitem{gaussian/kernel/cvpr18}
Mehdi S.~M. Sajjadi, Raviteja Vemulapalli, and Matthew Brown.
\newblock Frame-recurrent video super-resolution.
\newblock In {\em CVPR}, pages 6626--6634, 2018.

\bibitem{DBLP:conf/cvpr/ShaharFI11}
Oded Shahar, Alon Faktor, and Michal Irani.
\newblock Space-time super-resolution from a single video.
\newblock In {\em CVPR}, pages 3353--3360, 2011.

\bibitem{shan/sr/sa08}
Qi Shan, Zhaorong Li, Jiaya Jia, and Chi{-}Keung Tang.
\newblock Fast image/video upsampling.
\newblock {\em {ACM} TOG}, 27(5):153:1--153:7, 2008.

\bibitem{Irani/cvpr18}
Assaf Shocher, Nadav Cohen, and Michal Irani.
\newblock ``zero-shot" super-resolution using deep internal learning.
\newblock In {\em CVPR}, pages 3118--3126, 2018.

\bibitem{iGAN/kernel/nips19}
Assaf Shocher, Nadav Cohen, and Michal Irani.
\newblock Blind super-resolution kernel estimation using an internal-{GAN}.
\newblock In {\em NeurIPS}, 2019.

\bibitem{pwcnent/deqing}
Deqing Sun, Xiaodong Yang, Ming-Yu Liu, and Jan Kautz.
\newblock {PWC-Net}: {CNNs} for optical flow using pyramid, warping, and cost
  volume.
\newblock In {\em CVPR}, pages 8934--8943, 2018.

\bibitem{xintao/iccv17}
Xin Tao, Hongyun Gao, Renjie Liao, Jue Wang, and Jiaya Jia.
\newblock Detail-revealing deep video super-resolution.
\newblock In {\em ICCV}, pages 4482--4490, 2017.

\bibitem{TDAN}
Yapeng Tian, Yulun Zhang, Yun Fu, and Chenliang Xu.
\newblock {TDAN:} temporally deformable alignment network for video
  super-resolution.
\newblock {\em CoRR}, abs/1812.02898, 2018.

\bibitem{blindsr/sa09}
Cornill{\'{e}}re Victor, Djelouah Abdelaziz, Yifan Wang, Sorkine-Hornung Olga,
  and Schroers Christopher.
\newblock Blind image super-resolution with spatially variant degradations.
\newblock In {\em SIGGRAPH Asia}, 2019.

\bibitem{edvr}
Xintao Wang, Kelvin~C.K. Chan, Ke Yu, Chao Dong, and Chen Change~Loy.
\newblock {EDVR}: Video restoration with enhanced deformable convolutional
  networks.
\newblock In {\em CVPR Workshops}, 2019.

\bibitem{tof}
Tianfan Xue, Baian Chen, Jiajun Wu, Donglai Wei, and William~T Freeman.
\newblock Video enhancement with task-oriented flow.
\newblock {\em IJCV}, 127(8):1106--1125, 2019.

\bibitem{jiawei/cvpr17/deblur}
Jiawei Zhang, Jinshan Pan, Wei-Sheng Lai, Rynson W.~H. Lau, and Ming-Hsuan
  Yang.
\newblock Learning fully convolutional networks for iterative non-blind
  deconvolution.
\newblock In {\em CVPR}, pages 6969--6977, 2017.

\bibitem{kaizhang/cvpr17/deblur}
Kai Zhang, Wangmeng Zuo, Shuhang Gu, and Lei Zhang.
\newblock Learning deep {CNN} denoiser prior for image restoration.
\newblock In {\em CVPR}, pages 2808--2817, 2017.

\bibitem{zhangkai/cvpr19/sr}
Kai Zhang, Wangmeng Zuo, and Lei Zhang.
\newblock Deep plug-and-play super-resolution for arbitrary blur kernels.
\newblock In {\em CVPR}, pages 1671--1681, 2019.

\bibitem{RCAN}
Yulun Zhang, Kunpeng Li, Kai Li, Lichen Wang, Bineng Zhong, and Yun Fu.
\newblock Image super-resolution using very deep residual channel attention
  networks.
\newblock In {\em ECCV}, pages 294--310, 2018.

\end{thebibliography}
}

\end{document}